\documentclass[letterpaper]{article} 
\usepackage{aaai23}  
\usepackage{times}  
\usepackage{helvet}  
\usepackage{courier}  
\usepackage[hyphens]{url}  
\usepackage{graphicx} 
\usepackage{natbib}  
\usepackage{caption} 
\usepackage{algorithm}

\usepackage{soul}
\usepackage{caption}
\usepackage{amsmath}
\usepackage{amsthm}
\usepackage{color}
\usepackage{dblfloatfix}
\usepackage{balance}
\usepackage{amssymb}
\usepackage{booktabs}
\usepackage[noend]{algpseudocode}

\usepackage{subcaption}
\usepackage{bm}
\usepackage{multirow}
\usepackage{enumitem}
\newcommand{\eg}{{\it e.g.}}
\newcommand{\etal}{{\it et al.}}
\newcommand{\ie}{{\it i.e.}}

\newcommand{\wrt}{w.r.t. }
\newcommand{\stitle}[1]{\vspace{1mm} \noindent {\bf #1}}
\renewcommand{\vec}[1]{\ensuremath{\mathbf{#1}}}

\newcommand{\bG}{\ensuremath{\mathcal{G}}}
\newcommand{\bC}{\ensuremath{\mathcal{C}}}
\newcommand{\bV}{\ensuremath{\mathcal{V}}}
\newcommand{\bE}{\ensuremath{\mathcal{E}}}
\newcommand{\bD}{\ensuremath{\mathcal{D}}}

\newcommand{\bL}{\ensuremath{\mathcal{L}}}
\newcommand{\bN}{\ensuremath{\mathcal{N}}}
\newcommand{\bS}{\ensuremath{\mathcal{S}}}

\newcommand{\bY}{\ensuremath{\mathcal{Y}}}

\newcommand{\method}[1]{#1}
\newcommand{\model}{\method{DegFairGNN{}}}

\newcommand{\modelS}[1]{\method{DegFair{#1}}}

\newcommand{\eat}[1]{}

\newcommand\zeminC[1]{\begin{CJK*}{UTF8}{gbsn}\textcolor{blue}{[\textbf{Zemin:} #1]}\end{CJK*}}

\usepackage{newfloat}
\usepackage{listings}
\DeclareCaptionStyle{ruled}{labelfont=normalfont,labelsep=colon,strut=off} 
\floatstyle{ruled}
\newfloat{listing}{tb}{lst}{}
\floatname{listing}{Listing}
\title{On Generalized Degree Fairness in Graph Neural Networks}
\author{
    Zemin Liu, \textsuperscript{\rm 1}\thanks{Part of the work was done as a research scientist at Singapore Management University.}
    Trung-Kien Nguyen, \textsuperscript{\rm 2}
    Yuan Fang \textsuperscript{\rm 2}
}
\affiliations{
    \textsuperscript{\rm 1} National University of Singapore, Singapore \\
    \textsuperscript{\rm 2} Singapore Management University, Singapore

    zeminliu@nus.edu.sg, \{tknguyen, yfang\}@smu.edu.sg
}

\usepackage{bibentry}

\begin{document}

\maketitle

\begin{abstract}
Conventional graph neural networks (GNNs) are often confronted with fairness issues 
that may stem from their input, including node attributes and neighbors surrounding a node. 
While several recent approaches have been proposed to eliminate the bias rooted in sensitive attributes, 
they ignore the other key input of GNNs, namely the neighbors of a node, which can introduce bias since GNNs hinge on neighborhood structures to generate node representations. 
In particular, the varying neighborhood structures across nodes, manifesting themselves in drastically different node degrees, 
give rise to the diverse behaviors of nodes and biased outcomes. 
In this paper, we first define and generalize the degree bias using a generalized definition of node degree as a manifestation and quantification of different multi-hop structures around different nodes.
To address the bias in the context of node classification, we propose a novel GNN framework called Generalized Degree Fairness-centric Graph Neural Network (\model). 
Specifically, in each GNN layer, we employ a learnable debiasing function to generate debiasing contexts, which modulate the layer-wise neighborhood aggregation to eliminate the degree bias originating from the diverse degrees among nodes. 
Extensive experiments on three benchmark datasets demonstrate the effectiveness of our model on both accuracy and fairness metrics.
\end{abstract}

\section{Introduction}

Graph neural networks (GNNs) \cite{kipf2016semi,hamilton2017inductive,velivckovic2017graph} have emerged as a powerful family of graph representation learning approaches. 
They typically 
adopt a message passing framework, 
in which each node on the graph aggregates information from its neighbors recursively in multiple layers, unifying both node attributes and structures to generate node representations.
Despite their success,
GNNs are often confronted with fairness issues 
stemming from input node \emph{attributes} \cite{zemel2013learning,dwork2012fairness}. 
To be more specific, 
certain sensitive attributes (\eg, gender, age or race) may trigger bias and cause unfairness in downstream applications. Na\"ively removing the sensitive attributes does not adequately improve fairness, as the correlation between sensitive attributes and other attributes could still induce bias \cite{pedreshi2008discrimination}. 
Thus, prior  work on Euclidean data often employs additional regularizations or constraints to debias the model or post-processing the predictions \cite{hardt2016equality,pleiss2017fairness}. 
Similar principles have been extended to graphs
to eliminate bias rooted in sensitive node attributes \cite{rahman2019fairwalk,bose2019compositional,buyl2020debayes,li2020dyadic,dai2021say,fairview}.
For example, \method{CFC} \cite{bose2019compositional} tries to debias by applying some filters on the generated node representations, and some others resort to adversarial learning 
to 
detect sensitive attributes \cite{bose2019compositional,dai2021say}.

However, these approaches fail to account for the other key input of GNNs---the neighborhood structures of each node, 
which can also induce bias into node representations. 
Specifically, each node is linked to different neighbors, forming diverse neighborhood structures and manifesting drastically different node degrees. 
By virtue of the message passing framework, 
GNNs generate the representation of a node based on its neighborhood structures,
which is highly dependent on the abundance or scarcity of its neighboring nodes. 
Hence, drastic differences in node degrees could lead to differential node behaviors and biased outcomes.  
In particular, while the ``value'' of a node is usually co-determined by its own attributes and its neighboring structure (\ie, ``social capital''), the latter may systematically bias against newcomers despite their individual attributes, and established individuals may passively accumulate more social capital and become even harder for newcomers to compete.
For example, 
on a social network involving NBA players \cite{dai2021say}, a player with more followers may command a higher salary than another player with comparable physical ability and skills due to his popularity among fans; 
in a citation network, a paper with more initial citations may attract even more future citations than another paper with comparable quality on the same topic.
Generally, a node of larger degree is more likely to possess crucial advantages and thus receives more favorable outcomes than warranted.
This reveals a notable issue of \emph{degree fairness}, in which the degree-biased neighborhood structures can marginalize or even override the quality and competency of individual nodes.
The goal of degree fairness is to achieve equitable outcomes for nodes of different degrees, such that individuals of comparable competency should receive similar outcomes. More generally, we can also consider the degree bias stemming from multi-hop structures surrounding a node, using a \emph{generalized} definition of node degrees.

In this paper, we investigate the important problem of \emph{degree fairness in graph neural networks} in the context of node classification, which has not been explored to date.
On one hand, simply employing neighborhood sampling cannot address this issue due to the potential correlation between node attributes and their structures, and may further result in information loss \cite{wu2019net,tang2020investigating,liu2021tail}. 
On the other hand, node degree is a manifestation of neighborhood structures, which is fundamentally different from a sensitive node attribute \cite{rahman2019fairwalk,bose2019compositional,li2020dyadic,dai2021say}.
Particularly, in GNNs each node receives information from its neighbors, and thus nodes of diverse degrees can access varying amount of information, giving rise to the degree bias in learnt representations.
Moreover, the bias may be further amplified by multi-layered recursive neighborhood aggregation in GNNs.
While existing debiasing approaches that directly filter node representations are plausible for attribute-oriented fairness, they do not debias the core operation of neighborhood aggregation and thus cannot fundamentally alleviate the degree bias intrinsic to the diverse neighborhood structures. 

Toward degree fairness, we first introduce generalized degree to quantify the abundance or scarcity  of multi-hop contextual structures for each node, since in a broader sense the degree bias of a node stems not only from its one-hop neighborhood, but also its local context within a few hops.
We further propose a novel generalized \textbf{Deg}ree \textbf{Fair}ness-centric \textbf{GNN} framework named \model, which can work with any neighborhood aggregation-based GNN architecture. 
To fundamentally address degree fairness, 
we employ a learnable debiasing function to generate different \emph{debiasing contexts}, which are meant to balance 
the structural contrast between nodes with 
high and low generalized degrees. 
Specifically, the debiasing contexts directly target the neighborhood aggregation operation in each GNN layer, 
aiming to complement the neighborhood of low-degree nodes, while distilling the neighborhood of high-degree nodes.
As a result, 
the drastic differences in contextual structures across nodes can be balanced to reach a structurally fair state, enabling the generation of fair node representations.

To summarize, our contributions are three-fold. 
(1) We make the first attempt on defining and addressing generalized degree fairness in GNNs. 
(2) We propose a novel generalized degree fairness-centric GNN framework named \model\ that can flexibly work with neighborhood aggregation-based GNNs, to eliminate the generalized degree bias rooted in the layer-wise neighborhood aggregation. 
(3) Extensive experiments demonstrate that our proposed model is effective in both fairness and accuracy metrics.

\begin{figure*}[t]
\centering
\includegraphics[width=1.0\linewidth]{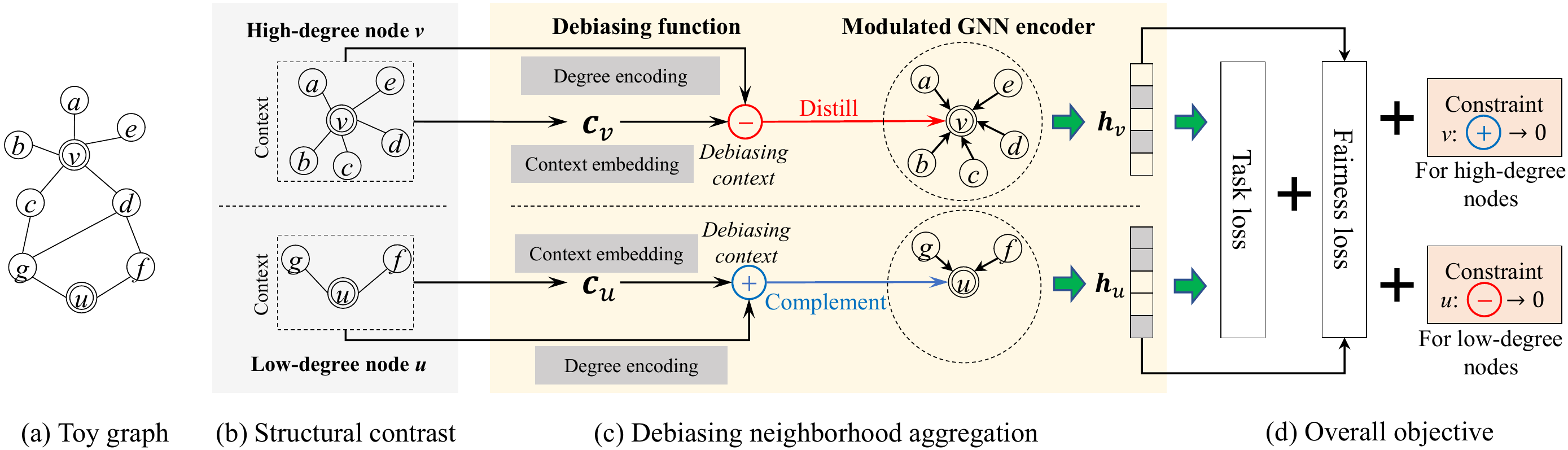}
\vspace{-3mm}
\caption{Overall framework of \model. 
}
\label{fig.framework}
\vspace{-3mm} 
\end{figure*}

\section{Problem Formulation} \label{sec.problem}
In this section, we introduce the definition of generalized degree fairness on graphs and several related concepts.

\stitle{Graph.} 
A graph is given by $\bG=\{\bV, \bE, \vec{X}\}$, where $\bV$ is the set of nodes, $\bE$ is the set of edges, and $\vec{X}\in\mathbb{R}^{|\bV|\times d_X}$ is the node attribute matrix with $d_X$ equalling the number of attributes. Equivalently, $\vec{x}_v\in\mathbb{R}^{d_X}$ is the attribute vector of node $v$. Let $\vec{A}\in\mathbb{R}^{|\bV|\times|\bV|}$ denote the corresponding adjacency matrix of $\bG$. 

\stitle{Generalized degree.}
In a broader sense, the degree bias on a node stems not only from its (one-hop) neighborhood, but also its local context within a certain number of hops.
Formally, the $r$-hop \emph{local context} of node $v$ is defined as 
$\bC_r=\{ v' \in \bV \mid d(v,v') \le r \}$, \ie, the set of nodes reachable from $v$ in at most $r$ steps, where $d(\cdot,\cdot)$ is the shortest distance from $v$ to $v'$ on the graph.
Subsequently, we introduce \emph{generalized degree} to quantify the abundance of multi-hop local context surrounding a node.
Given a node $v$, we define its $r$-hop generalized degree $\deg_r(v)\in\mathbb{R}$ as
\begin{equation}\label{eq.context-struct}
\deg_r(v) = [\vec{A}^{r}\mathbf{1}]_v,
\end{equation}
where  $r\in\mathbb{N}^+$ is the number of hops, 
$\mathbf{1}\in\mathbb{R}^{|\bV|}$ is a column vector filled with ones, and $[\vec{x}]_i$ denotes the $i$-th entry in vector $\vec{x}$.
Here $[\vec{A}^r]_{i,j}$, the $(i,j)$ entry in the matrix $\vec{A}^r$, gives the number of walks of length $r$ from node $i$ to node $j$. 
Thus, $\deg_r(v)$ represents the total number of (non-distinct) nodes reachable from $v$ in $r$ hops, a straightforward measure of the abundance of $v$'s $r$-hop local context. 
As a special case, $\deg_1(v)$ is simply the (1-hop) degree of $v$.

\stitle{Generalized degree fairness.}
Generalized degree fairness amounts to node representations that are free from variant generalized degrees.
In the context of multi-class node classification, 
we formalize two fairness definitions in terms of statistical parity \cite{dwork2012fairness} and equal opportunity \cite{hardt2016equality}. Both definitions rely on the notion of generalized degree groups, in which we divide the nodes into $m$ mutually exclusive groups $\bG_1,\ldots,\bG_m$ so that 
\begin{align}
    \bG_i = \{v\in \bV \mid d_i \le \deg_r(v) < d_{i+1}\},
\end{align}
where $d_1<d_2<\ldots<d_{m+1}$ are a series of degree boundaries between the groups, and $d_1=\min_{v\in\bV} \deg_r(v)$ and $d_{m+1}=\max_{v\in\bV}\deg_r(v)+1$.
Furthermore, let $\bY$ be the set of classes. For ease of discussion, hereafter we may simply call ``generalized degree'' as ``degree''.

To achieve \emph{Degree Statistical Parity} (DSP), we require the class predictions to be independent of node degree groups. 
That is, for any class $y\in\bY$ and any two groups $\bG_i$ and $\bG_j$, 
\begin{equation}
 P(\hat{y}_v=y|v \in\bG_i) = P(\hat{y}_v=y|v \in \bG_j),
\end{equation}
where $\hat{y}_v$ is the predicted class of $v$.
To achieve \emph{Degree Equal Opportunity} (DEO), we require the probability of each node being predicted into its true class $y$ to be equal for nodes in different degree groups. That is, 
\begin{align}
   P(\hat{y}_v=y|y_v = y,v \in\bG_i) \vspace{-4mm}
 = \vspace{-4mm} P(\hat{y}_v=y|y_v = y,v \in \bG_j),  \nonumber
\end{align}
where $y_v$ is the true class of $v$.

Our degree fairness is defined in a group-wise manner, which is in line with existing fairness definitions \cite{zafar2017fairness,agarwal2021towards} and offers a flexible setup. On the one hand, a simple and practical scenario could involve just two groups, since typically the degree fairness issue is the most serious between nodes with the smallest and largest degrees. On the other hand, the most fine-grained groups would be each unique value of degree  forming its own group, but it is not ideal for two reasons. First, some groups may have very few nodes depending on the degree distribution, leading to large variance. Second, the smoothness of degrees are lost: nodes of similar degrees are expected to have rather similar advantage or disadvantage compared to nodes of very different degrees.

\section{Proposed Model: \model}

In this section, we introduce the proposed \model, starting with some preliminaries on GNNs.

\subsection{Preliminaries on GNNs}

Modern  GNNs usually resort to multi-layer neighborhood aggregation, in which each node recursively aggregates information from its neighbors. Specifically, in the $l$-th layer  the representation of node $v$, $\vec{h}^l_v\in \mathbb{R}^{d_l}$, is constructed as 
\begin{equation}\label{eq:gnn}
    \vec{h}^l_v = \sigma\left(\textsc{Aggr}\left(\{\vec{h}^{l-1}_u\mid u\in\bN_v\};\omega^l\right)\right),
\end{equation}
where $d_l$ is the dimension of node representations in the $l$-th layer, $\textsc{Aggr}(\cdot)$ denotes an aggregation function such as mean-pooling \cite{kipf2016semi} or self-attention \cite{velivckovic2017graph}, $\sigma$ is an activation function, $\bN_v$ denotes the set of neighbors of $v$, and $\omega^l$ denotes the learnable parameters in layer $l$.
Node representations in the input layer are given by the input node attributes, \ie, $\vec{h}^0_v \equiv \vec{x}_v$.

\subsection{Overall Framework}


The overall framework of \model\ is illustrated in Fig.~\ref{fig.framework}. We first design a strategy of \emph{structural contrast} in Fig.~\ref{fig.framework}(b), to split nodes into two groups by their degree. The two groups contrast each other to facilitate the learning of a \emph{debiasing function}, which generates node-specific \emph{debiasing contexts} to modulate neighborhood aggregation, as shown in Fig.~\ref{fig.framework}(c). More specifically, the modulation aims to distill high-degree nodes to remove bias, and complement low-degree nodes to enrich their degree-biased neighborhood structures. Thus, by trying to balance the two groups, the modulation is able to debias neighborhood aggregation. Finally, the debiasing function and GNN are jointly optimized by the task loss and the fairness loss in Fig.~\ref{fig.framework}(d).

\subsection{Structural Contrast} \label{sec.stru-contrast}

To debias nodes with varying contextual structures, we propose to learn the debiasing function by contrasting between two groups of nodes, namely, low-degree nodes $\bS_0$ and high-degree nodes $\bS_1$, as illustrated in Fig.~\ref{fig.framework}(b). 
Note that in the layer-wise neighborhood aggregation in Eq.~\eqref{eq:gnn}, each node can only access its one-hop local contexts in each layer. 
Thus, it is a natural choice to construct the two groups based on one-hop degree: 
$\bS_0=\{v\in\bV\mid\deg_1(v)\leqslant K\}$ and $\bS_1=\bV \setminus \bS_0$.
That is, $\bS_0$ is the low-degree group, $\bS_1$ is the high-degree group, and $K$ is a threshold hyperparameter. 
Note that having two groups can provide the most significant contrast for model training.

To contrast between the two groups, nodes in the same group would share the learnable parameters for the debiasing mechanism, whereas nodes across different groups would have different parameters. 
This strategy enables the two groups to eliminate the degree bias in different ways, which is desirable given their different degrees. 

\subsection{Debiasing Neighborhood Aggregation} 

To fundamentally eliminate degree bias, we propose to debias neighborhood aggregation, the key operation in GNNs in which degree bias is rooted. As shown in Fig.~\ref{fig.framework}(c), we leverage a learnable debiasing function to generate the debiasing contexts, which modulate the neighborhood aggregation for each node and in each layer of the GNN encoder.

\stitle{Debiasing function.}
More concretely, we leverage a debiasing function $\bD(\cdot)$ with learnable parameters to fit each group, due to the divergence between the two groups in terms of node degree.
On one hand, for a low-degree node $u\in \bS_0$, $\bD(u;\theta^l_0)\in \mathbb{R}^{d_l}$ outputs the debiasing context for $u$  in layer $l$ to complement the neighborhood of $u$. 
On the other hand, for a high-degree node $v \in \bS_1$, $\bD(v;\theta^l_1)\in \mathbb{R}^{d_l}$ outputs the debiasing context for $v$  in layer $l$ to distill the neighborhood of $v$.
Note that each group $\bS_*$ has its own parameters $\theta^l_*$ for $\bD(\cdot)$ in layer $l$.
The learning is guided by a \emph{fairness loss} in the overall objective (see Sect.~\ref{sec.loss}), which drives the debiasing contexts toward distilling information on the high-degree nodes while complementing information on the low-degree nodes, in order to achieve the balance and fairness between the two groups.

The debiasing function is designed to generate a debiasing context for each node in each layer,
to modulate the neighborhood aggregation in a node- and layer-wise manner.
To achieve ideal modulations, the debiasing contexts should be \emph{comprehensive} and \emph{adaptive}.

\textit{\textbf{Comprehensiveness.}} First, debiasing contexts need to be comprehensive, to account for both the content and structure information in the neighborhood. 
To be comprehensive, we resort to the \emph{context embedding} of node $v$ in layer $l$, denoted $\vec{c}^{l}_v\in\mathbb{R}^{d_{l-1}}$, which can be calculated as 
\begin{equation} \label{eq.context-emb}
    \vec{c}^{l}_v=\textsc{Pool}(\{\vec{h}^{l-1}_u \mid u \in \bC_r(v)\}),
\end{equation}
where $\textsc{Pool}(\cdot)$ is a pooling function. Here we use a simple mean pooling, although it can also be made learnable. 
The context embedding $\vec{c}^{l}_v$ aggregates the layer-($l$-$1$) contents in node $v$'s $r$-hop local context  $\bC_r(v)$, and thus embodies both the content and structure information. The debiasing function is then a function of the context embedding, \ie, 
\begin{align}
\bD(v;\theta^l_*)=f(\vec{c}^{l}_v;\theta^l_{c,*}),
\end{align}
which is parameterized by $\theta^l_{c,*}$, where $*$ is $0$ or $1$ depending on the node group. We use a fully connected layer as $f$.

\textit{\textbf{Adaptiveness.}}  Second, debiasing contexts need to be \emph{adaptive}, to sufficiently customize to each node.
While the two groups of nodes, $\bS_0$ and $\bS_1$, are already differentiated by group-specific parameters for the debiasing function, the differentiation is coarse-grained and cannot sufficiently adapt to individual nodes. In particular, even for nodes in the same group, their degrees still vary, motivating the need for finer-grained node-wise adaptation.
However, letting each node have node-specific parameters is not feasible due to a much larger model size, which tends to cause overfitting and scalability issues.
For fine-grained adaptation without blowing up the model size, inspired by hypernetworks \cite{ha2016hypernetworks,perez2018film,liu2021nodewise}, we adopt a secondary neural network to generate node-wise transformations on the debiasing contexts in each layer. Concretely, the debiasing function can be reformulated as
\begin{equation} \label{eq:full-debias-function}
    \bD(v;\theta^l_*)=(\gamma^{l}_v+\vec{1})\odot f(\vec{c}^{l}_v;\theta^l_{c,*})+\beta^{l}_v, 
\end{equation}
where $\gamma^{l}_v$ and $\beta^{l}_v\in\mathbb{R}^{d_{l}}$ are node-specific scaling and shifting operators in layer $l$, respectively. Here $\odot$ denotes element-wise multiplication, and \vec{1} is a vector of ones to ensure that the scaling is centered around one. Note that $\gamma^{l}_v$ and $\beta^{l}_v\in\mathbb{R}^{d_{l}}$ are not directly learnable, but are respectively generated by a secondary network conditioned on each node's degree. Specifically, 
\begin{align} \label{eq.film-factors}
    \gamma^{l}_v = \phi_\gamma(\delta^l(v);\theta^l_{\gamma}), \quad \beta^{l}_v = \phi_\beta(\delta^l(v);\theta^l_{\beta}),
\end{align}
where $\phi_\gamma$ and $\phi_\beta$ can be any neural network, and we simply use a fully connected layer.
The input to these secondary networks, $\delta^l(v)$, is the \emph{degree encoding} of $v$ to condition the transformations on the degree of $v$, which we will elaborate later. 
Thus, the learnable parameters of the debiasing function in Eq.~\eqref{eq:full-debias-function} become $\theta^l_*=\{\theta^l_{c,*},\theta^l_{\gamma},\theta^l_{\beta}\}$ in layer $l$, and $\theta_*=\{\theta^1_*, \theta^2_*, \ldots \}$ denotes the full set of learnable parameters of the debiasing function for nodes in group $\bS_*$.

Finally, the input to the secondary network in Eq.~\eqref{eq.film-factors} is the degree encoding $\delta^l(v)$, instead of the degree of $v$ itself. 
The reason is that degree is an ordinal variable, implicating that degree-conditioned functions should be smooth w.r.t.~small changes in degree. 
In other words, nodes with similar degrees should undergo similar transformations, and vice versa. In light of this, inspired by positional encoding \cite{vaswani2017attention}, we propose to encode the degree of a node $v$ in layer $l$ into a vector
$\delta^l(v)\in\mathbb{R}^{d_l}$, such that
    $[\delta^l(v)]_{2i}=\sin(\deg_1(v)/10000^{2i/d_l})$ and $[\delta^l(v)]_{2i+1}=\cos(\deg_1(v)/10000^{2i/d_l})$.
Note that, although nodes at the periphery of the two groups $\bS_0$ and $\bS_1$ (\eg, nodes with degrees $K$ and $K+1$) have different debiasing parameters ($\theta_{c,0}$ and $\theta_{c,1}$), the usage of degree encoding in Eqs.~\eqref{eq:full-debias-function} and \eqref{eq.film-factors} will assimilate their debiasing contexts to some extent given their similar degrees.

\stitle{Modulated GNN encoder.}
Given the debiasing contexts, we couple it with the standard neighborhood aggregation in Eq.~\eqref{eq:gnn}, and modulate neighborhood aggregation for the two groups differently. Specifically, in layer $l$,
\begin{align} \label{eq.aggr}
&\vec{h}^l_v = \sigma\bigg(\textsc{Aggr}(\{\vec{h}^{l-1}_u\mid u\in\bN_v\};\omega^l) \nonumber\\[-2mm]
&+ \epsilon\cdot \big(\underbrace{I(v\in \bS_0)  \bD(v;\theta^l_{0})}_{\text{complement low-deg.~group}} + \underbrace{I(v\in \bS_1)  \bD(v;\theta^l_{1})}_{\text{distill high-deg.~group}}\big)\bigg),
\end{align}
where $I(\cdot)$ is a 0/1 indicator function based on the truth value of its argument, and $\epsilon>0$ is a hyperparameter to control the impact of the debiasing contexts.

\subsection{Training Constraints and Objective}\label{sec.loss}

We focus on the task of node classification. 
Apart from the classification loss and the fairness loss associated with the classification, we further introduce several useful constraints to improve the training process, as shown in Fig.~\ref{fig.framework}(d). 

\stitle{Classification loss.}
For node classification, the last layer of the GNN typically sets the dimension to the number of classes, and uses a softmax activation such that the $i$-th output dimension is the probability of class $i$. With a total of $\ell$ layers, 
the output node representations $\{\vec{h}^{\ell}_v: v\in \bV^{\text{tr}}\}$ of the training nodes $\bV^{\text{tr}}$ can be fed into the cross-entropy loss, 
\begin{align}\label{eq:loss}
    \bL_{1} = -&\textstyle\sum_{v\in \bV^{\text{tr}}}\sum_{y=1}^{|\bY|} [\vec{y}_{v}]_y \ln [\vec{h}^{\ell}_{v}]_y,
\end{align}
where $\vec{y}_v$ is the one-hot vector of $v$'s class.

\stitle{Fairness loss.}
As the proposed debiasing function in Eq.~\eqref{eq:full-debias-function} is learnable, it should be driven toward fair representations.
Thus, we further employ a fairness loss on the training data.  Let $\bS_0^\text{tr}$ and $\bS_1^\text{tr}$ denote the group of low- and high-degree nodes in the training data, respectively. We use a DSP-based loss, trying to achieve parity in the predicted probabilities for the two groups, as follows.  
\begin{equation} \label{eq:fairness-loss} 
    \bL_2 = \textstyle\left\|{\textstyle\frac{1}{|\bS_0^\text{tr}|}}\sum_{v\in \bS_0^\text{tr}}\vec{h}^{\ell}_v-{\textstyle\frac{1}{|\bS_1^{\text{tr}}|}}\sum_{v\in \bS_1^{\text{tr}}}\vec{h}^{\ell}_v\right\|^2_2,
\end{equation}
where $\|\cdot\|^2_2$ is the $L_2$ norm.
This fairness loss drives the learnable debiasing function toward the fairness metric (\eg, DSP here) to guide the training of \model, by constraining the debiasing function to learn how to distill information on high-degree nodes and complement information on low-degree nodes, respectively.
Besides, it is also possible to apply DEO. 
Note that, 
the fairness loss aims to constrain the prediction distribution to be similar across the two groups, yet does not require the node representations to be similar. Thus, \method{\model} would not worsen the over-smoothing phenomenon \cite{pmlr-v80-xu18c}.

\stitle{Constraints on debiasing contexts.}
For a low-degree node $u$, its debiasing context $\bD(u;\theta^l_{0})$ aims to complement but not distill its neighborhood.
On the contrary, for a high-degree node $v$, its debiasing context $\bD(v;\theta^l_{1})$ aims to distill but not complement its neighborhood.
Thus, both $\bD(\cdot;\theta^l_{1})$ for low-degree nodes in $\bS_0$ and $\bD(\cdot;\theta^l_{0})$ for high-degree nodes in $\bS_1$ should be close to zero.
The two constraints promote the learning of debiasing contexts by contrasting between the two groups, which can be formulated as the following loss.
\begin{align} \label{eq.debias-context-constraint}
    \bL_3 =  \sum_{l=1}^{\ell}\bigg(\sum_{v\in\bS_0^{\text{tr}}} \|\bD(v;\theta^l_{1})\|^2_2 
     + \sum_{v\in\bS_1^{\text{tr}}}\|\bD(v;\theta^l_{0})\|^2_2\bigg).
\end{align}

\stitle{Constraints on scaling and shifting.}
To prevent overfitting the data with arbitrarily large scaling and shifting, we further consider the loss below to restrict the search space. 
\begin{equation} \label{eq.film-factors-constraints}
    \bL_4=\textstyle\sum_{l=1}^{\ell}\sum_{v\in\bV^{\text{tr}}}(\|\gamma^{l}_v\|^2_2 + \|\beta^{l}_v\|^2_2).
\end{equation}



\stitle{Overall loss.}
By combining all the above loss terms, we formulate the overall loss as
\begin{equation} \label{eq.overall-obj}
    \bL = \bL_1 + \mu \bL_2 + \lambda (\bL_3 + \bL_4), 
\end{equation}
where 
$\mu,\lambda$ are hyper-parameters. 
In Appendix~A, we outline the training procedure and give a complexity analysis. 

\section{Experiments} \label{sec.experiments}

In this section, we evaluate the proposed \model
in terms of both accuracy and fairness.

\subsection{Experimental Setup} \label{sec.expe-setup}

\stitle{Datasets.}
We use two Wikipedia networks, \emph{Chameleon} and \emph{Squirrel} \cite{pei2019geom}, in which each node represents a Wikipedia  page, and each edge denotes a reference between two pages in either direction. 
We split the nodes into five categories \wrt their traffic volume for  classification. 
We also use a citation network \emph{EMNLP} \cite{ma2020copulagnn}, in which each node denotes a paper published in the  conference, and each edge denotes two papers that have been co-cited. 
We split the nodes into two categories \wrt their out-of-EMNLP citation count for classification. 
We summarize the datasets in Table~\ref{table.datasets}, and present more details in Appendix~B.
Note that, both traffic volumes and citation counts can be deemed as individual benefits of the nodes that tend to be biased by the node degree, motivating us to employ these datasets. 
In these contexts, a GNN that predicts the benefit outcome independently of degree groups is trying to be fair, which should focus on the relevance and quality of the nodes rather than their existing links.


\begin{table}[!t]
\centering
\small
\caption{Summary of datasets.\label{table.datasets}}
\vspace{-2mm}
\addtolength{\tabcolsep}{0pt}
\scalebox{0.95}{
\begin{tabular}{@{}c|rrrc@{}}
\toprule
 Dataset	&    Nodes 	& Edges 	&   Features  &  Classes \\
\midrule
 Chameleon   & 2,277  &  31,371  &    2,325  & 5 (traffic volume)     \\
 Squirrel   & 5,201  &  198,353  &    2,089  & 5 (traffic volume)      \\ 
 EMNLP  & 2,600  &  7,969  &    8  & 2 (citation count)       \\\bottomrule
\end{tabular}}
\vspace{-4mm}
\end{table}

\stitle{Base GNNs.}
Our proposed \model\ can work with different GNN backbones. 
We employ \method{GCN} \cite{kipf2016semi} as the default base GNN in our experiments, and name the corresponding fairness model \method{\modelS{GCN}}.
We also adopt two other base GNNs, \method{GAT} \cite{velivckovic2017graph} and \method{GraphSAGE} \cite{hamilton2017inductive},
as described in Appendix~C. 

\stitle{Baselines.}
We consider the following two categories of
baselines.
(1) \emph{Degree-specific models}: \method{DSGCN} \cite{tang2020investigating}, \method{Residual2Vec} \cite{kojaku2021residual2vec} and Tail-GNN \cite{liu2021tail}. They employ degree-specific operations on the nodes \wrt their degrees, 
to improve task accuracy especially for the low-degree nodes.
(2) \emph{Fairness-aware models}: \method{FairWalk} \cite{rahman2019fairwalk}, \method{CFC} \cite{bose2019compositional},  \method{FairGNN} \cite{dai2021say}, \method{FairAdj} \cite{li2020dyadic} and \method{FairVGNN} \cite{fairview}. They are proposed to address the sensitive attribute-based fairness of nodes on graphs.
%
To apply them to degree fairness, we define the generalized node degree as the sensitive attribute. 
However, 
even so,
these fairness-aware baselines do not debias the neighborhood aggregation mechanism in each GNN layer, and thus fall short of addressing the degree bias from the root. 
More details of the baselines are given in Appendix~D.

\stitle{Data split and parameters.} 
For all the datasets, we randomly split the nodes into 
training, validation and test set
with proportion 6:2:2. We set the threshold $K$ for the structural contrast in Sect.~\ref{sec.stru-contrast} 
as the mean node degree by default.
We further analyze the impact of $K$ on both accuracy and fairness in Appendix~G.2.
For other hyper-parameter settings, please refer to Appendix~E.


\begin{table*}[!t]
    \centering
    \small
   \caption{Comparison with baselines ($r=1$, 20\% Top/Bottom). 
    } \label{table.baselines-20-1-layer}
    \vspace{-2mm} 
    {\scriptsize Henceforth, tabular results are in percent with standard deviation over 5 runs; the best fairness result is \textbf{bolded} and the runner-up is \underline{underlined}.\vspace{1mm}}
    \addtolength{\tabcolsep}{-1mm}
    \resizebox{1\linewidth}{!}{
    \begin{tabular}{@{}c|c||c|ccc|ccccc|c@{}}
    \toprule
     \multicolumn{2}{c||}{} & \method{GCN}  & \method{DSGCN} & \method{Residual2Vec} & \method{Tail-GNN} & \method{FairWalk} & \method{CFC} & \method{FairGNN} & \method{FairAdj} & \method{FairVGNN} & \method{\modelS{GCN}}  \\ \midrule\midrule
     \multirow{3}{*}{{\centering Chamel.}} & Acc. $\uparrow$ & 62.45 $\pm$ 0.21 & 63.90 $\pm$ 1.28  & 49.04 $\pm$ 0.01 & 66.08 $\pm$ 0.19 & 56.36 $\pm$ 0.75 & 63.02 $\pm$ 0.84 & 70.70 $\pm$ 0.52 & 51.71 $\pm$ 1.13 & 72.32 $\pm$ 0.50  & 69.91 $\pm$ 0.19 \\
     & $\Delta_{\text{DSP}}\downarrow$ & \ \ 9.68 $\pm$ 1.37  & \ \ 8.81 $\pm$ 1.15  & 14.52 $\pm$ 0.69 & \ \ 8.51 $\pm$ 1.72 & \ \ 8.18 $\pm$ 0.93  & 10.12 $\pm$ 1.28 & \ \ \underline{7.33} $\pm$ 1.09  & \ \ 9.79 $\pm$ 1.91  & \ \ 8.86 $\pm$ 1.11  & \ \ \textbf{5.85} $\pm$ 0.32  \\
     & $\Delta_{\text{DEO}}\downarrow$ & 36.08 $\pm$ 2.65 & 25.14 $\pm$ 2.67 & 37.31 $\pm$ 1.99 & 26.09 $\pm$ 3.25 & \underline{22.89} $\pm$ 2.75 & 29.54 $\pm$ 1.95 & 26.83 $\pm$ 1.95 & 27.48 $\pm$ 2.06 & 26.02 $\pm$ 2.39 & \textbf{21.60} $\pm$ 0.71  \\ \midrule
     
     \multirow{3}{*}{{\centering Squirrel}} & Acc. $\uparrow$ & 47.85 $\pm$ 1.33 & 40.71 $\pm$ 2.17 & 28.47 $\pm$ 0.01 & 42.62 $\pm$ 0.06 & 37.68 $\pm$ 0.65 & 45.64 $\pm$ 2.19 & 57.29 $\pm$ 0.77 & 35.18 $\pm$ 1.22 & 46.97 $\pm$ 0.48 & 59.21 $\pm$ 0.97  \\
     & $\Delta_{\text{DSP}}\downarrow$ & 13.37 $\pm$ 2.83 & 16.08 $\pm$ 0.86 & 25.11 $\pm$ 0.48 & 18.91 $\pm$ 0.26 & \ \ \textbf{7.94} $\pm$ 0.36  & 12.40 $\pm$ 0.48 & 12.96 $\pm$ 1.03 & 16.63 $\pm$ 1.56 & 26.67 $\pm$ 0.52 & \ \ \underline{9.54} $\pm$ 1.02  \\
     & $\Delta_{\text{DEO}}\downarrow$ & 27.00 $\pm$ 3.79 & 32.61 $\pm$ 3.74 & 34.49 $\pm$ 0.72 & 33.60 $\pm$ 0.72 & \underline{17.12} $\pm$ 1.50 & 21.60 $\pm$ 2.69 & 17.62 $\pm$ 2.40 & 27.54 $\pm$ 1.73 & 35.80 $\pm$ 1.76 & \textbf{16.42} $\pm$ 1.38  \\ \midrule
     
     \multirow{3}{*}{{\centering EMNLP}} & Acc. $\uparrow$ & 78.92 $\pm$ 0.43 & 82.19 $\pm$ 0.77 & 80.69 $\pm$ 0.01 & 83.72 $\pm$ 0.28 & 82.23 $\pm$ 0.18 & 80.15 $\pm$ 1.13 & 86.81 $\pm$ 0.22 & 76.50 $\pm$ 1.55 & 84.03 $\pm$ 0.34 & 79.92 $\pm$ 0.77  \\
     & $\Delta_{\text{DSP}}\downarrow$ & 44.55 $\pm$ 1.90 & 50.00 $\pm$ 2.98 & \underline{12.90} $\pm$ 0.15 & 41.18 $\pm$ 1.58 & 33.52 $\pm$ 1.46 & 56.60 $\pm$ 1.95 & 58.23 $\pm$ 1.44 & 40.38 $\pm$ 4.64 & 43.92 $\pm$ 1.43 & \textbf{12.38} $\pm$ 3.72 \\
     & $\Delta_{\text{DEO}}\downarrow$ & 34.05 $\pm$ 3.56 & 46.92 $\pm$ 2.91 & \underline{11.26} $\pm$ 0.67 & 36.76 $\pm$ 1.48 & 30.67 $\pm$ 1.42 & 45.21 $\pm$ 2.27 & 51.56 $\pm$ 1.38 & 41.89 $\pm$ 4.78 & 40.95 $\pm$ 1.71 & \ \ \textbf{8.52} $\pm$ 2.26  \\\bottomrule
     \end{tabular}} 
     \vspace{-3mm}
\end{table*}

\stitle{Evaluation.}
For model \emph{performance}, we evaluate the node classification accuracy on the test set.

For model \emph{fairness}, 
the group-wise DSP and DEO (Sect.~\ref{sec.problem}) essentially require that the outcomes are independent of the degree groups.
In particular, we form two groups from the test set: $\bG_0$ and $\bG_1$, containing test nodes with low and high generalized degrees, respectively. 
A two-group granule is a first major step toward degree fairness (resource-poor group {\it vs}.~resource-rich group), where the fairness issue is the most serious.
As degree usually follows a long-tailed distribution \cite{liu2020towards}, 
based on the Pareto principle 
\cite{newman2005power} 
we select the top 20\% test nodes by generalized degree as $\bG_1$, and the bottom 20\% nodes as $\bG_0$, which may present more prominent biases.
To thoroughly evaluate the fairness, we also test on alternative groups with top and bottom 30\% nodes. 
%
Moreover, generalized degree fairness is defined \wrt a pre-determined number of hops $r$.
We employ $r=1$ 
as the default, and further report the evaluation with $r=2$.
We only select $r\le 2$ for evaluation as GNNs usually have shallow layers, which implies a small $r$ can sufficiently cover the contextual structures where biases are rooted.
Hence, 
we adopt the following metrics $\Delta_{\text{DSP}}$ and $\Delta_{\text{DEO}}$, which evaluate the mean difference between the distributions of the two groups ($\bG_1$ and $\bG_0$) in the test set.
For both metrics, a smaller value implies a better fairness.
\begin{align}
\Delta_{\text{DSP}} = \textstyle\frac{1}{|\bY|}\sum_{y\in\bY}  \big|& P(\hat{y}_v=y|v\in \bG_0)\ - \nonumber\\
&  P(\hat{y}_v=y|v\in \bG_1)\big|,
\\
\Delta_{\text{DEO}} = \textstyle\frac{1}{|\bY|}\sum_{y\in\bY}  \big|& P(\hat{y}_v=y|y_v=y,v\in \bG_0)\ - \nonumber\\
& P(\hat{y}_v=y|y_v=y,v\in \bG_1)\big|.
\end{align}

\subsection{Model Accuracy and Fairness} \label{sec.eva}



We first evaluate our model using the default base GNN (\ie, GCN) and fairness settings, and further supplement it with additional fairness settings and base GNNs.

\stitle{Main evaluation.}
Using default fairness settings, we first evaluate \method{\modelS{GCN}} (\ie, on a GCN backbone) against the baselines in Table~\ref{table.baselines-20-1-layer}, and 
make the following observations.

Firstly, \method{\modelS{GCN}} consistently outperforms the baselines in both fairness metrics, while achieving a comparable classification accuracy, noting that there usually exists a trade-off between fairness and accuracy \cite{bose2019compositional,dai2021say}. 
The only exception is on the Squirrel dataset, where \method{FairWalk} obtains the best $\Delta_{\text{DSP}}$ at the expense of a significant degradation in accuracy, while \method{\modelS{GCN}} remains a close runner-up ahead of other baselines.
Secondly, degree-specific models \method{DSGCN}, \method{Residual2Vec} and \method{Tail-GNN} typically have worse fairness than fairness-aware models, since they exploit the structural difference between nodes for model performance, rather than to eliminate the degree bias for model fairness.
Thirdly, the advantage of \method{\modelS{GCN}} over existing fairness-aware models including \method{FairWalk}, \method{CFC}, \method{FairGNN}, \method{FairAdj} and \method{FairVGNN} implies that degree fairness needs to be addressed at the root, by debiasing the core operation of layer-wise neighborhood aggregation. 
Merely treating degree as a sensitive attribute 
cannot fundamentally alleviate the structural degree bias. Interestingly, \method{GCN} can generally achieve comparable and sometimes even better fairness than these methods, which again shows that degree fairness cannot be sufficiently addressed without directly debiasing  neighborhood aggregation.

\begin{table}[!t]
    \centering
   \small
   \caption{Comparison to baselines ($r=2$, 20\% Top/Bottom).
    } \label{table.baselines-20-2-layer}
    \vspace{-2mm}
    \addtolength{\tabcolsep}{-1mm}
    \resizebox{1.0\linewidth}{!}{
    \begin{tabular}{@{}c|c||c|cc|c@{}} 
    \toprule
     \multicolumn{2}{c||}{} & \method{GCN} & \method{FairWalk} & \method{FairGNN} & \method{\modelS{GCN}}  \\ \midrule\midrule
     \multirow{3}{*}{{\centering Chamel.}} & Acc. $\uparrow$ & 62.45 $\pm$\ 0.21 & 56.36 $\pm$\ 0.75 & 70.70 $\pm$\ 0.52 & 69.91 $\pm$\ 0.19   \\ 
     & $\Delta_{\text{DSP}}\downarrow$ & \ \ \underline{5.96} $\pm$ 0.89 & 10.38 $\pm$ 0.85 & \ \ 6.70 $\pm$ 0.32  & \ \ \textbf{5.25} $\pm$ 0.39  \\
     & $\Delta_{\text{DEO}}\downarrow$ & 26.92 $\pm$ 2.09  & 25.46 $\pm$ 1.66 & \underline{23.66} $\pm$ 0.93 & \textbf{19.05} $\pm$ 0.74 \\ \midrule
    
    \multirow{3}{*}{{\centering Squirrel}} & Acc. $\uparrow$ & 47.85 $\pm$\ 1.33 & 37.68 $\pm$\ 0.65 & 57.29 $\pm$\ 0.77 & 59.21 $\pm$\ 0.97   \\ 
    & $\Delta_{\text{DSP}}\downarrow$ & 14.61 $\pm$\ 2.63 & \ \ \underline{9.64} $\pm$\ 0.50 & 11.11 $\pm$\ 0.93 & \ \ \textbf{8.26} $\pm$\ 0.57  \\ 
     & $\Delta_{\text{DEO}}\downarrow$ & 28.62 $\pm$\ 3.89 & 17.37 $\pm$\ 1.10     & \underline{16.29} $\pm$\ 2.07 & \textbf{14.95} $\pm$\ 1.22 \\ \midrule
    
     \multirow{3}{*}{{\centering EMNLP}} & Acc. $\uparrow$ & 78.92 $\pm$\ 0.43 & 82.23 $\pm$\ 0.18 & 86.81 $\pm$\ 0.22 & 79.92 $\pm$\ 0.77   \\ 
     & $\Delta_{\text{DSP}}\downarrow$ & 45.03 $\pm$ 1.77 & \underline{34.80} $\pm$ 1.26 & 52.88 $\pm$ 1.39 & \textbf{10.87} $\pm$ 4.00 \\
     & $\Delta_{\text{DEO}}\downarrow$ & 34.71 $\pm$ 3.31 & \underline{31.11} $\pm$ 1.34 & 45.78 $\pm$ 1.36 & \ \ \textbf{8.72} $\pm$ 2.17 \\\bottomrule
     \end{tabular}}
     \vspace{-2mm}
\end{table}

\begin{table}[!t]
    \centering
   \small
   \caption{Comparison to baselines ($r=1$, 30\% Top/Bottom).
   } \label{table.baselines-30-1-layer}
    \vspace{-2mm}
    \addtolength{\tabcolsep}{-1mm}
    \resizebox{1.0\linewidth}{!}{
    \begin{tabular}{@{}c|c||c|cc|c@{}} 
    \toprule
     \multicolumn{2}{c||}{} & \method{GCN} & \method{FairWalk} & \method{FairGNN} & \method{\modelS{GCN}}  \\ \midrule\midrule
     \multirow{3}{*}{{\centering Chamel.}} & Acc. $\uparrow$ & 62.45 $\pm$\ 0.21 & 56.36 $\pm$\ 0.75 & 70.70 $\pm$\ 0.52 & 69.91 $\pm$\ 0.19   \\  
     & $\Delta_{\text{DSP}}\downarrow$ & \ \  \underline{5.95} $\pm$ 1.02 & \ \  8.16 $\pm$ 0.38  & \ \  6.92 $\pm$ 0.29 & \ \  \textbf{4.15} $\pm$ 0.02 \\
     & $\Delta_{\text{DEO}}\downarrow$ & 18.00 $\pm$ 1.76 & 16.65 $\pm$ 1.32 & \underline{14.52} $\pm$ 1.09 & \ \ \textbf{8.39} $\pm$ 0.37 \\ \midrule
    
    \multirow{3}{*}{{\centering Squirrel}} & Acc. $\uparrow$ & 47.85 $\pm$\ 1.33 & 37.68 $\pm$\ 0.65 & 57.29 $\pm$\ 0.77 & 59.21 $\pm$\ 0.97   \\ 
    & $\Delta_{\text{DSP}}\downarrow$ & 10.34 $\pm$\ 2.15 & \ \ \textbf{6.17} $\pm$\ 0.36 & \ \ 9.27 $\pm$\ 0.68 & \ \ \underline{7.39} $\pm$\ 0.63 \\
     & $\Delta_{\text{DEO}}\downarrow$ & 22.62 $\pm$\ 3.10 & \textbf{14.97} $\pm$\ 1.12 & \underline{17.42} $\pm$\ 1.11 & 17.71 $\pm$\ 1.05 \\ \midrule
     
     \multirow{3}{*}{{\centering EMNLP}} & Acc. $\uparrow$ & 78.92 $\pm$\ 0.43 & 82.23 $\pm$\ 0.18 & 86.81 $\pm$\ 0.22 & 79.92 $\pm$\ 0.77   \\ 
     & $\Delta_{\text{DSP}}\downarrow$ & 42.87 $\pm$ 1.40 & \underline{34.19} $\pm$ 0.91 & 48.25 $\pm$ 1.97 & \textbf{14.46} $\pm$ 3.35 \\
     & $\Delta_{\text{DEO}}\downarrow$ & 37.89 $\pm$ 3.27 & \underline{34.49} $\pm$ 0.91 & 48.83 $\pm$ 1.97 & \textbf{10.92} $\pm$ 2.87 \\\bottomrule
     \end{tabular}}
     \vspace{-3mm}
\end{table}

\stitle{Additional fairness settings.}
We further evaluate fairness using different test groups, \ie, $r=2$ with 20\% top/bottom in Table~\ref{table.baselines-20-2-layer} and $r=1$ with 30\% top/bottom in Table~\ref{table.baselines-30-1-layer}. Note that, accuracy evaluation is applied on the whole test set regardless of the groups, 
so the accuracies are identical to those reported in Table~\ref{table.baselines-20-1-layer}. Here, we compare 
with two representative baselines,  
and observe that, even with different test groups, our proposed \method{\model} can generally outperform the baselines in terms of degree fairness.
Similar to Table~\ref{table.baselines-20-1-layer}, \method{FairWalk} can perform better in fairness metrics in Table~\ref{table.baselines-30-1-layer} at the expense of significantly worse accuracy.

\stitle{Additional base GNNs.}
In addition to the default base GCN, we further experiment with other base GNNs, namely, GAT and GraphSAGE, and form two new models, \ie, \modelS{GAT} and \modelS{SAGE}, respectively.
Their performance under the default fairness settings is reported in Table~\ref{table.base-models}, while the other settings (\ie, $r=2$, 20\% Top/Bottom, and $r=1$, 30\% Top/Bottom) can be found in Appendix~G.1.

Altogether, the three base GNNs employ different neighborhood aggregations, \eg, mean or attention-based mechanisms, but none of them employs a fairness-aware aggregation. 
The results show that our models can generally outperform their corresponding base GNN models across the three datasets in terms of fairness, demonstrating the flexibility of \model\ when working with different GNN backbones.

\begin{table}[!t]
    \centering
    \small
    \caption{With other base GNNs ($r=1$, 20\% Top/Bottom).
    } \label{table.base-models}
    \addtolength{\tabcolsep}{-0.4mm}
    \vspace{-2mm}
    \resizebox{1.0\linewidth}{!}{
    \begin{tabular}{@{}c|c||cc|cc@{}} 
    \toprule
     \multicolumn{2}{c||}{} & \method{GAT} & \method{\modelS{GAT}} & \method{GraphSAGE} & \method{\modelS{SAGE}}  \\ \midrule\midrule
     
     \multirow{3}{*}{{\centering Chamel.}} & Acc. $\uparrow$ & 63.15 $\pm$ 0.40 & 69.64 $\pm$ 0.44 & 53.15 $\pm$ 0.56 & 60.95 $\pm$ 0.84 \\
     & $\Delta_{\text{DSP}}\downarrow$ & \ \ 9.35 $\pm$ 1.61  & \ \ \textbf{7.88} $\pm$ 1.30  &  10.86 $\pm$ 0.74 & \ \ \textbf{8.22} $\pm$ 1.22  \\
     & $\Delta_{\text{DEO}}\downarrow$ & 29.59 $\pm$ 1.43 & \textbf{26.12} $\pm$ 2.06 & 29.42 $\pm$ 1.57 & \textbf{26.40} $\pm$ 2.32 \\ \midrule
     
     \multirow{3}{*}{{\centering Squirrel}} & Acc. $\uparrow$ & 41.44 $\pm$ 0.21 & 45.55 $\pm$ 1.44 & 34.39 $\pm$ 0.62 & 34.63 $\pm$ 1.31  \\
     & $\Delta_{\text{DSP}}\downarrow$ & 12.60 $\pm$ 0.77 & \textbf{12.03} $\pm$ 0.63 & \ \ 5.39 $\pm$ 0.66  & \ \ \textbf{3.76} $\pm$ 0.23  \\
     & $\Delta_{\text{DEO}}\downarrow$ & 24.89 $\pm$ 0.69 & \textbf{20.64} $\pm$ 3.06 & 17.13 $\pm$ 2.86 & \textbf{14.91} $\pm$ 1.35 \\ \midrule
     
     \multirow{3}{*}{{\centering EMNLP}} & Acc. $\uparrow$ & 70.42 $\pm$ 0.77 & 81.57 $\pm$ 1.14 & 83.96 $\pm$ 0.31 & 83.57 $\pm$ 0.44 \\
     & $\Delta_{\text{DSP}}\downarrow$ & 24.40 $\pm$ 3.06 & \textbf{14.11} $\pm$ 6.28 & 56.33 $\pm$ 1.12 & \textbf{28.43} $\pm$ 3.79 \\
     & $\Delta_{\text{DEO}}\downarrow$ & \ \ \textbf{8.36} $\pm$ 1.29  & 12.28 $\pm$ 6.19 & 51.71 $\pm$ 0.88 & \textbf{24.65} $\pm$ 3.35 \\\bottomrule
    \end{tabular}}
     \vspace{-1mm}
\end{table}

\subsection{Model Analysis} \label{sec.analysis}
We conduct further model analysis on \model, using the default base GNN and fairness settings. 


\stitle{Ablation Study.} \label{sec.ablation}
To validate the contribution of each module in \modelS{GCN}, we consider several variants.
(1) \emph{no scale \& shift}: we remove the scaling and shifting operations for the debiasing contexts, \ie, no node-wise adaptation is involved.
(2) \emph{no contrast}: we remove the structural contrast, and 
utilize a common debiasing function with shared parameters for all the nodes.
(3) \emph{no modulation}: we remove the modulation (complementing or distilling) from Eq.~\eqref{eq.aggr}, resulting in a standard \method{GCN} equipped with an additional fairness loss.

\begin{figure}[t]
\centering
\includegraphics[width=1\linewidth]{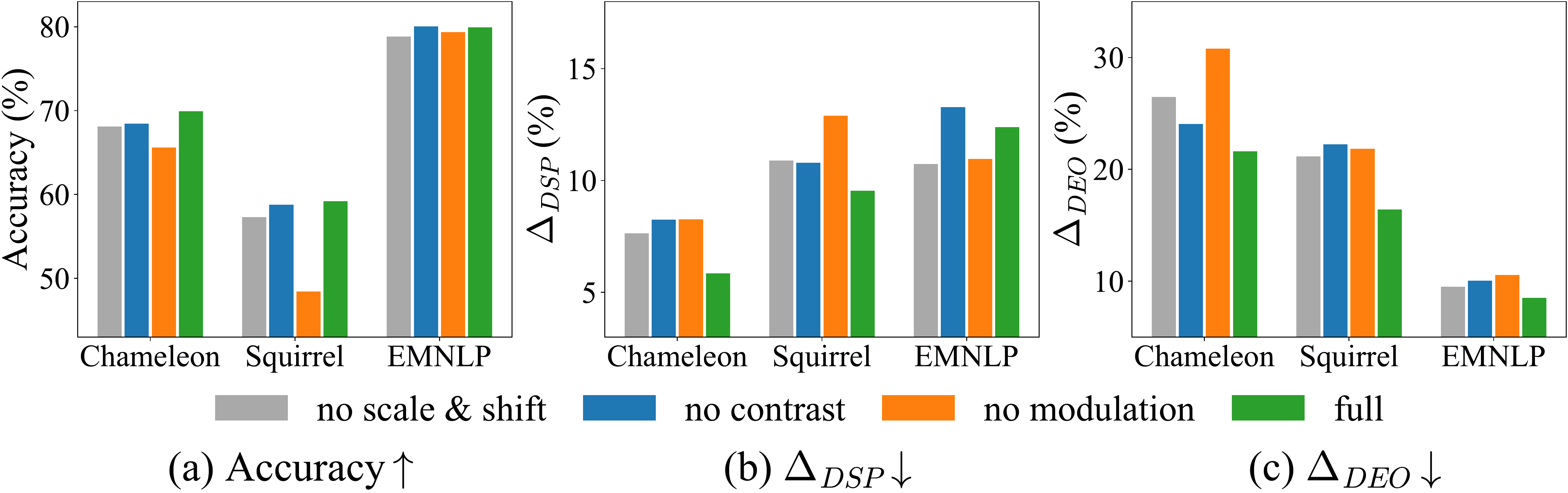} 
\vspace{-4mm}
\caption{Ablation study on the effect of each module.}
\label{fig.ablation}
  \vspace{-3mm}
\end{figure}

We report the results in Fig.~\ref{fig.ablation} and make several observations.
Firstly, without scaling and shifting, we often get worse accuracy and fairness, which means node-wise adaptation is useful and the model can benefit from the finer-grained treatment of nodes. 
Secondly, without structural contrast, the fairness metrics generally become worse.
Thus, structural contrast is effective in driving the debiasing contexts toward degree fairness. 
Lastly, without the modulation of neighborhood aggregation, the fairness metrics become worse in most cases, implying that simply adding a fairness loss without properly debiasing the layer-wise neighborhood aggregation does not work well.

\stitle{Other analyses.}
We present further analyses on the threshold $K$, scalability and parameter sensitivity in Appendix G.2, G.3, and G.4, respectively, due to space limitation.

\section{Related Work}

We only present the most related work here, while leaving the rest to Appendix H due to space limitation.


\stitle{Fairness learning.}
Fairness learning \cite{zemel2013learning,kusner2017counterfactual} can be broadly categorized into three kinds.
(1) Pre-processing methods usually eliminate bias by reshaping the dataset \cite{feldman2015certifying}, such as correcting the labels \cite{kamiran2009classifying}.
(2) In-processing methods usually rely on model refinement for debiasing, such as applying additional regularizations or constraints  \cite{dwork2012fairness,zafar2017fairness}.
(3) Post-processing methods \cite{hardt2016equality,pleiss2017fairness} usually designate new labels on the predictions to remove bias.

Some recent approaches \cite{rahman2019fairwalk,bose2019compositional,dai2021say,fairview} deal with the sensitive attribute-based fairness on graphs.
\method{FairWalk} \cite{rahman2019fairwalk} tries to sample fair paths by guiding random walks based on sensitive attributes, while
\method{FairAdj} \cite{li2020dyadic} studies dyadic fairness for link prediction by adjusting the adjacency matrix.
On the other hand, 
EDITS \cite{dong2022edits} proposes to debias the input attributed data so that GNNs can be fed with less biased data, while
FairVGNN \cite{fairview} tries to address the issue of sensitive attribute leakage by automatically identifying and masking correlated attributes.
Others \cite{bose2019compositional,dai2021say,feng2019learning} employ discriminators \cite{goodfellow2014generative} as additional constraints on the encoder to facilitate the identification of sensitive attributes. 
\method{DeBayes} \cite{buyl2020debayes} trains a conditional network embedding 
\cite{kang2019conditional} by using a biased prior and evaluates the model with an oblivious prior, thus reducing the impact of sensitive attributes.
Ma \etal\ \cite{ma2021subgroup} investigate the performance disparity between test groups rooted in the distance between them and the training instances.
Dong \etal\ \cite{dong2021individual} study the individual fairness for GNNs from the ranking perspective.
Besides, there are fairness learning studies on heterogeneous \cite{zeng2021fair} and knowledge graphs \cite{fisher2020debiasing}.
However, none of these works is designed for degree fairness on graphs.


\stitle{Degree-specific GNNs.}
Some recent studies investigate the influence of degrees on the performance of GNNs. Various strategies have been proposed, such as employing degree-specific transformations on nodes \cite{wu2019net,tang2020investigating}, accounting for their different roles \cite{ahmed2020role}, balancing the sampling of nodes in random walks 
\cite{kojaku2021residual2vec}, or considering the degree-related performance differences between nodes \cite{kang2022rawlsgcn,liu2020towards,liu2021tail}. 
However, they emphasize the structural difference between nodes in order to improve task accuracy, rather than to eliminate the degree bias for fairness.

\section{Conclusions}

In this paper, we investigated the important problem of degree fairness on GNNs. In particular, we made the first attempt on defining and addressing the generalized degree fairness issue. To eliminate the degree bias rooted in the layer-wise neighborhood aggregation, we proposed a novel generalized degree fairness-centric GNN framework named \model, which can flexibly work with most modern GNNs. The key insight is to target the root of degree bias, by modulating the core operation of neighborhood aggregation through a structural contrast. We conducted extensive experiments on three benchmark datasets and achieved promising results on both accuracy and fairness metrics. 


\section{Acknowledgments}
This research is supported by the Agency for Science, Technology
and Research (A*STAR) under its AME Programmatic Funds (Grant
No. A20H6b0151).

\bibliography{references}

\end{document}